\documentclass[11pt,a4paper]{article}
\usepackage[hyperref]{acl2019}
\usepackage{times}
\usepackage{latexsym}
\usepackage{{graphicx}}
\usepackage{multirow}
\usepackage{url}
\usepackage{pbox}

\aclfinalcopy 

\title{Emotional Embeddings: Refining Word Embeddings to Capture Emotional Content of Words}

\author{Armin Seyeditabari \\
	UNC Charlotte  \\
\small	\texttt{sseyedi1@uncc.edu} \normalsize \\\And
	Narges Tabari \\
	University of Virginia \\
\small	\texttt{ns5kn@virginia.edu} \normalsize \\\And
	Shefie Gholizade \\
	UNC Charlotte  \\
\small	\texttt{sgholiza@uncc.eduu} \normalsize \\\And
	Wlodek Zadrozny \\
	UNC Charlotte \\
\small	\texttt{wzadrozn@uncc.edu} \normalsize }
\date{}

\begin{document}
\maketitle
\begin{abstract}
Word embeddings are one of the most useful tools in any modern natural language processing expert's toolkit. They contain various types of information about each word which makes them the best way to represent the terms in any NLP task. But there are some types of information that cannot be learned by these models. Emotional information of words are one of those. In this paper, we present an approach to incorporate emotional information of words into these models. We accomplish this by adding a secondary training stage which uses an emotional lexicon and a psychological model of basic emotions. We show that fitting an emotional model into pre-trained word vectors can increase the performance of these models in emotional similarity metrics. Retrained models perform better than their original counterparts from 13\% improvement for Word2Vec model, to 29\% for GloVe vectors. This is the first such model presented in the literature, and although preliminary, these emotion sensitive models can open the way to increase performance in variety of emotion detection techniques.

\end{abstract}

\section{Introduction}

There is an abundant volume of textual data available online about variety of subjects through social media. This availability of large amount of data led to a fast growth in information extraction using natural language processing. One of the most important types of information that can be captured is the affective reaction of the population to a specific event, product, etc. We have seen a vast improvement in extracting the sentiment from text to the point that sentiment analysis has become one of the standard tools in any NLP expert's toolkit and has been used in various applications\citep{ravi2015survey}. 

On the other hand, emotion detection, as a more fine-grained affective information extraction technique, is just recently making larger appearance in the literature. The amount of useful information which can be gained by moving past the negative and positive sentiments and towards identifying discrete emotions can help improve many applications. For example, the two emotions \textit{Fear} and \textit{Anger} both express negative opinion of a person toward something, however, it has been shown that fearful people tend to have pessimistic view of the future, while angry people tend to have more optimistic view \citep{lerner2000beyond}. Moreover, fear generally is a passive emotion, while anger is more likely to lead to action \citep{miller2009relative}. The usefulness of understanding emotions in political science \citep{Druckman2008}, psychology, marketing \citep{bagozzi1999role}, human-computer interaction \citep{brave2003emotion}, and many more, gave the field of emotion detection in natural language processing life of its own, resulting in a surge of research papers in recent years. 
%

Word embeddings, as one of the best methods to create representation for each word in the corpus, is mostly used as features in any neural network base classifiers. These word vectors are created in manner that the angular distance between them represents various types of information. For example, the distance between the two words \textit{cat} and \textit{feline} should be less that the distance between \textit{cat} and \textit{canine} as cat is a feline but not a canine. 

You can find verity of similarity, or categorical information in the shape of these vector spaces that make them one of the best tools we have in natural language processing. But these embeddings, due to the nature of their training methods, do not contain the emotional similarity information. In this paper, we present and analyze a methodology to incorporate emotional information into these models after the fact. We accomplish this by utilizing an emotion model and an emotion lexicon -in this case Plutchik's wheel of emotions  \citep{plutchik1991emotions}, and NRC emotion lexicon \citep{mohammad2013crowdsourcing}. We have also used a secondary emotion model to create an emotional similarity test to compare the performance of the models before and after training.

This preliminary result is an important step to show the potential that these models can be used to improve emotion detection systems in different ways. Emotion sensitive embeddings can be used in various emotion detection methodologies, such as recurrent neural network classifiers to possibly improve the model performance in learning, and classifying emotions. It can also be used in attention networks \citep{yang2016hierarchical} to calculate feature weights for each term in the corpus to potentially improve the classification accuracy by giving more weights to the emotionally charged terms.

\begin{figure}[t]
	\centering
	\includegraphics[width=1.0\linewidth]{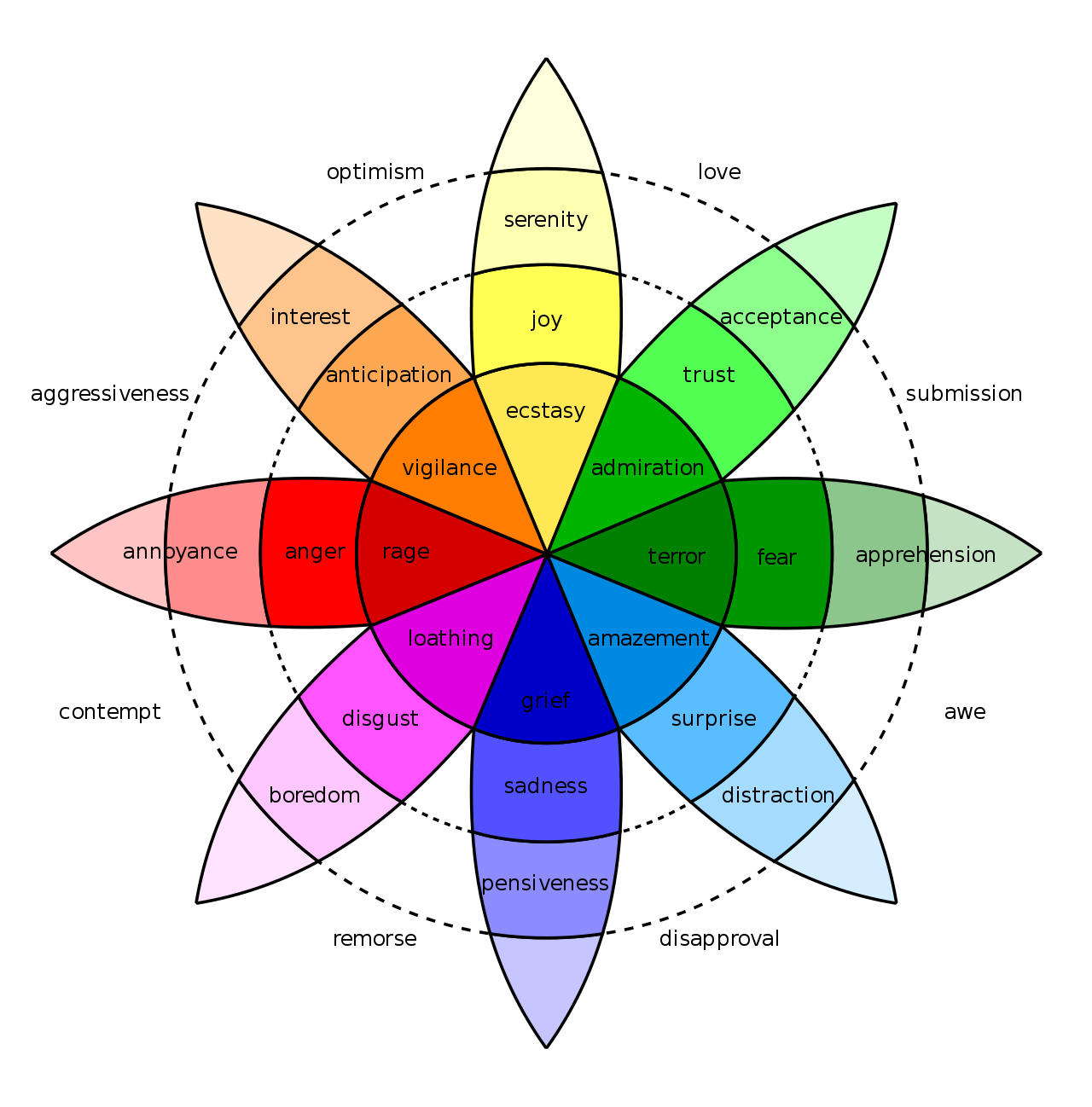}
	\caption{Plutchik's Wheel of Emotions. Opposite emotions are placed on opposite  petals.}
	\label{fig:1280px-plutchik-wheel}
\end{figure}

\section{Related Work}

In the past decade, specially by increasing usage of neural networks, word embeddings have been one of the most useful tools in natural language processing. Word2Vec created by Mikolov et al., and presented in two papers \citeyearpar{mikolov2013distributed,mikolov2013efficient} has shown that these vectors could perform reliably in variety of tasks. GloVe \citep{pennington2014glove} took a different approach for creating word embeddings which performed on par with Word2Vec. 

After the success of these models, many studies have been done to figure out the shortcomings of these models and try to make them better. Speer et al. used an ensemble method to integrate Word2Vec and GloVe with ConceptNet knowledge base \citep{speer2016conceptnet} and created ConceptNet NumberBatch model \citep{speer2016ensemble} and showed that their model outperform either of those models in verity of tasks. \citet{faruqui2014retrofitting} also presented a method to refine these vectors base on an external semantic lexicon by encouraging vectors for similar words move closer to each other. 

Understanding that these embedding models do not perform well for semantically opposite words, in their paper \citet{mrkvsic2016counter}  created a methodology that not only brought the vectors for similar words close to each other, but also moved vectors for opposite words farther apart. \citet{mikolov2017advances} created an improved model fastText in which they  used combination of known tricks to make the vectors perform better in different tasks. But as shown in \citet{seyeditabari2017can} these models do not perform well in emotional similarity tasks.

There has also been various attempts to create sentiment embeddings that would perform better in sentiment analysis tasks compared to standard vector spaces \citep{tang2014learning,tang2016sentiment,yu2017refining}. Moving past sentiments, in this paper, we present a method to incorporate emotional information of words into some of these models mentioned above. 


\section{Fitting Emotional Constraints in Word Vectors}

For fitting emotional information into pre-trained word vectors, we use a methodology similar to what \citet{mrkvsic2016counter} used to incorporate additional linguistic constraints in word vector spaces. Our goal here, is to change the vector space \(V=\{v_1, v_2, \ldots, v_n\}\) to \(V'=\{v'_1, v'_2, \ldots, v'_n\}\) in a careful manner to add emotional constraints to the vector space without loosing too much information already present during the original learning step. To preform this task we create two sets of constraints based on NRC emotion lexicon, one for words which have positive relation to an emotion such as (abduction, sadness), and one to keep track of each words relation to the opposite of that emotion (abduction, joy), joy being the opposite of sadness. In NRC lexicon Mohammad et al. annotated over 14k English words for eight emotions from Plutchik’s model of basic emotions(See Figure \ref{fig:1280px-plutchik-wheel}).

\begin{table*}[t!]
	\centering
	\tiny
	\begin{tabular}{|l|l|l|}
		\hline
		
		\textbf{Primary Emotion} & \textbf{Secondary Emotion} & \textbf{Tertiary Emotion}\\
		\hline
		\multirow{3}{*}{Liking} & Affection & \pbox{250pt}{Adoration · Fondness · Liking · Attractiveness ·  Caring · Tenderness · Compassion · Sentimentality}\\
		& Lust/Sexual desire& Desire · Passion · Infatuation\\
		& Longing & Longing\\
		\hline
		
		\multirow{7}{*}{Joy} & Cheerfulness & \pbox{250pt}{Amusement · Bliss · Gaiety · Glee · Jolliness · Joviality · Joy · De-light · Enjoyment · Gladness · Happiness · Jubilation · Elation · Satisfaction · Ecstasy · Euphoria}\\
		& Zest& Enthusiasm · Zeal · Excitement · Thrill · Exhilaration\\
		& Contentment & Pleasure\\
		& Pride & Triumph\\
		& Optimism & Eagerness · Hope\\
		& Enthrallment & Enthrallment · Rapture\\
		& Relief & Relief\\
		\hline
		
		\multirow{1}{*}{Surprise} & Surprise & Amazement · Astonishment\\
		\hline
		
		\multirow{6}{*}{Anger} & Irritability & Aggravation · Agitation · Annoyance · Grouchy · Grumpy · Crosspatch\\
		& Exasperation& Frustration\\
		& Rage & \pbox{250pt}{Anger · Outrage · Fury · Wrath · Hostility · Ferocity · Bitter · Hatred · Scorn · Spite · Vengefulness · Dislike · Resentment}\\
		& Disgust & Revulsion · Contempt · Loathing\\
		& Envy & Jealousy\\
		& Torment & Torment\\
		\hline
		
		\multirow{6}{*}{Sadness} & Suffering & Agony · Anguish · Hurt\\
		& Sadness& \pbox{250pt}{Depression · Despair · Gloom · Glumness · Unhappy · Grief · Sor-row · Woe · Misery · Melancholy}\\
		& Disappointment & Dismay · Displeasure\\
		& Shame & Guilt · Regret · Remorse\\
		& Neglect & \pbox{250pt}{Alienation · Defeatism · Dejection · Embarrassment · Homesickness · Humiliation · Insecurity · Insult · Isolation · Loneliness · Rejection}\\
		& Sympathy & Pity · Mono no aware · Sympathy\\
		\hline
		
		\multirow{2}{*}{Fear} & Horror & \pbox{250pt}{Alarm · Shock · Fear · Fright · Horror · Terror · Panic · Hysteria · Mortification}\\
		& Nervousness&  \pbox{250pt}{Anxiety · Suspense · Uneasiness · Apprehension (fear) · Worry · Distress · Dread}\\
		\hline
		
	\end{tabular}
	\caption{Three layered emotion classification.
	}
	\label{tab:parrots}
\end{table*}
\normalsize

 To create our two constraint sets, we extract all word/emotion relations indicated in the lexicon so that in our first set \(S = \{(w_1,e_1),(w_1,e_3), (w_2, e_2), \ldots \}\) we have ordered pairs, each indicating a word and the emotion it is associated with. And for each emotion  \(e_i\), we add its opposite  \(e'_i\)  to our second set \(O=\{(w_1,e'_1),(w_1,e'_3), (w_2, e'_2), \ldots\}\) in which \(e'_i\) is the opposite emotion to \(e_i\) based on Plutchik’s model. We have extracted over 8k such pairs of (word, emotion) constraints from NRC lexicon for each of the positive and negative relation sets.

We define our objective functions so that we decrease the angular distance between words with their associated emotion in the set S, and at the same time, increase their distance with their opposite emotions in the set O. We want the pairs of words in positive relation set to get closer together, so the objective function for positive relations would be:
\begin{equation}
PR(V') = \sum_{(u,w)  \in S}^{} max(0, d(v'_u, v'_w))
\end{equation}

where \(d(v'_u, v'_w)\) is the cosine distance between the two vectors. And we want to increase the distance between pairs of words in our negative relation set, so the objective function for the negative relations would be:
\begin{equation}
NR(V') = \sum_{(u,w)  \in O}^{} max(0, 1 - d(v'_u, v'_w))
\end{equation}
\normalsize
We also need to make sure we lose as little information as possible by preserving the shape of our original vector space. In order to do this, we add a third part to our objective function to make sure we are not changing overall shape of the space by much:

\small
\begin{equation}
VSP(V,V') = \sum_{i=1}^{N}\sum_{j  \in N(i)}^{} max(0, |d(v'_u, v'_w) - d(v_u, v_w)|)
\end{equation}
\normalsize

For efficiency purposes, we only calculate the distance for a neighborhood of each word \(N(i)\) which includes all words within the radius distance \(r=0.2\) of the word. So our final objective function is the sum of all three together:

\small
\begin{equation}
Obj(V') =  PR(V') + NR(V') + VSP(V,V') 
\end{equation}
\normalsize

Stochastic gradient decent was used for 20 epochs to train the vector space \(V\) and generate the new space \(V'\). 

\begin{table*}[t!]
	\centering
	\small
	\begin{tabular}{cc}
		\begin{tabular}{ccccc}
			\hline
			& \multicolumn{2}{c}{\textbf{Sadness vs. Joy}} & \multicolumn{2}{c}{\textbf{Anger vs. Fear}} \\	
			&  \textbf{Before} & \textbf{After} &   \textbf{Before} & \textbf{After} \\\hline
			Word2Vec & 0.32 & 0.16 & 0.31 & 0.09\\
			GloVe & 0.23 & 0.11 & 0.19 & \textbf{0.04} \\
			fastText & 0.38 & 0.17 & 0.33 & 0.12\\
			Numberbatch & 0.23 &\textbf{ 0.10} & 0.19 & 0.05 \\
			\hline
		\end{tabular} 
	& 
		\begin{tabular}{cc}
			\hline
			\multicolumn{2}{c}{\textbf{In-category Similarity}}\\
			\textbf{Before} & \textbf{After}\\\hline
		0.45 & 0.51 \\
		0.38 & 0.49 \\
		0.44 & 0.50 \\
		0.47 & \textbf{0.57} \\
			\hline
		\end{tabular}
	\end{tabular}
	\caption{\textbf{Left}: Average similarity between opposite emotion groups. We want the similarity of opposite emotions be as close to zero az possible. After training the average similarities decrease for all models . \textbf{Right}: Average of in-category mutual similarity in three layered categorization of emotions before and after emotional fitting. We want the similarity of close emotions be as close to one as possible. After training, average similarity of in-category emotions increases for all models.}\label{tab:opposites}
\end{table*}

\section{Experiments}

In our experiment we compared variety of word embeddings with their emotionally fitted counterparts for various metrics based on emotional models. As we trained the model on Plutchik's model we decided to use another emotion model for testing. In the first experiment we assess the average in-category mutual similarity of  secondary and tertiary emotions in the three level categorization of emotions described by  \citet{shaver1987emotion}. In this model, Shaver et al. defined 6 basic emotions of \textit{Liking, Joy, Surprise, Anger, Sadness, and Fear}, and categorized around 140 sub-emotions under these 6 emotions in two layers (See Table \ref{tab:parrots}). The reported numbers are the average cosine similarity of all mutual in-category emotions words, and can be seen in Table \ref{tab:opposites}. The vector spaces used  here are:

\small
\begin{itemize}
	\item Word2Vec trained full English Wikipedia dump
	\item GloVe from their own website
	\item fastText  trained with subword information on Common Crawl
	\item ConceptNet Numberbatch
\end{itemize}
\normalsize

%
%

It is clear that each emotionally fitted vector space is preforming much better than its original counterpart from 13\% improvement for Word2Vec model, to 29\% for GloVe vectors. Overall best performance belongs to emotionally fitted ConceptNet Numberbatch by average similarity score of 0.57 up from 0.47 (See Table \ref{tab:opposites}).


In the second experiment we assessed the performance of the model for similarity between opposite emotions. Again we used Shaver et al.'s categorization as our testing emotion model and calculated the mutual similarity between opposite emotion groups. In this test we chose two pairs of opposite emotions, \textit{Joy} vs. \textit{Sadness} and \textit{Anger} vs. \textit{Fear}. The reported numbers are average cosine similarity between each member of the opposite emotion categories.


As shown in Table \ref{tab:opposites} the models perform significantly better after training with best performance for the retrained Numberbatch in distinguishing between \textit{Anger} vs. \textit{Fear} and retrained Glove for \textit{Joy} vs. \textit{Sadness} (with Numberbatch following closely). All emotionally fitted vector spaces can be accessed via \href{https://goo.gl/R43CEQ}{this link}\footnote{https://goo.gl/R43CEQ}.

\section{Conclusion and Future Work}

Embedding models have an important role in word representation in various natural language processing tasks. They are able to preserve many types information about the terms, but not emotional information which due to its complexity is hard to grasp just from the statistical information of the corpus. In this paper, we have proposed an approach to incorporate emotional information of words into these models in a second stage of training, and showed that this methodology is able to increase the performance of the embedding model in the defined emotional similarity metrics from 13\% increase for Word2Vec to 29\% increase for Glove with the best performance belonging to retrained Conceptnet Numberbatch. While this is the first steps toward creating and analyzing emotional embeddings, further study is required, and being done, to test various emotional models and lexicons, and to improve emotional information that can be incorporated into these models. 

This methodology could be improved by incorporating more complex emotional information, such as intensity of emotions and combination of emotions defined in psychological models. Training the original embeddings on corpora that are more emotionally rich might also increase the emotion sensitivity of these models. With the absence of established emotional similarity metrics, we chose a secondary emotion model to create a basic similarity test. With lack of standard metrics and corpora, further testing of the model is required to see how they increase performance in affect related NLP tasks such as detecting emotion in text using recurrent neural networks.

\bibliography{acl2019}
\bibliographystyle{acl_natbib}

\end{document}